\documentclass[conference]{IEEEtran}
\IEEEoverridecommandlockouts
\usepackage{cite}
\usepackage{amsmath,amssymb,amsfonts}
\usepackage{graphicx}
\usepackage{textcomp}
\usepackage{xcolor}
\def\BibTeX{{\rm B\kern-.05em{\sc i\kern-.025em b}\kern-.08em
    T\kern-.1667em\lower.7ex\hbox{E}\kern-.125emX}}

\usepackage{amsmath,amssymb,amsfonts,mathtools}
\usepackage{breqn}
\usepackage{algorithm2e}
\usepackage{algpseudocode}
\usepackage{adjustbox}
\usepackage{tfrupee}
\usepackage{tcolorbox}
\usepackage{tabularx}
\usepackage{array}
\usepackage{colortbl}
\usepackage{caption}
\tcbuselibrary{skins}
\usepackage{subfig}

\usepackage[utf8]{inputenc}
\usepackage[english]{babel}

\algnewcommand{\LeftComment}[1]{\Statex \(\triangleright\) #1}

\usepackage{url}
\usepackage[misc,geometry]{ifsym}

\begin{document}

\title{Enhancement to Training of Bidirectional GAN :  An Approach to Demystify Tax Fraud}

\author{\IEEEauthorblockN{1\textsuperscript{st} Priya Mehta}
\IEEEauthorblockA{\textit{Welingkar Institute of Management Development and Research} \\
Mumbai, India \\
priya.mehta@welingkar.org}

\and
\IEEEauthorblockN{2\textsuperscript{nd} Sandeep Kumar}
\IEEEauthorblockA{\textit{Department of Computer Science and Engineering} \\
\textit{ Indian Institute of Technology Hyderabad}\\
Sangareddy, India \\
sandeep@iith.ac.in}

\and
\IEEEauthorblockN{3\textsuperscript{rd} Ravi Kumar}
\IEEEauthorblockA{\textit{Department of Computer Science and Engineering} \\
\textit{Indian Institute of Technology Hyderabad}\\
Sangareddy, India \\
CS18MTECH11028@iith.ac.in}

\and
\IEEEauthorblockN{4\textsuperscript{th} Ch. Sobhan Babu}
\IEEEauthorblockA{\textit{Department of Computer Science and Engineering} \\
\textit{Indian Institute of Technology Hyderabad}\\
Sangareddy, India \\
sobhan@cse.iith.ac.in}
}

\maketitle

\begin{abstract}
Outlier detection is a challenging activity. Several machine learning techniques are proposed in the literature for outlier detection. In this article, we propose a new training approach for  bidirectional GAN (BiGAN) to detect outliers. To validate the proposed approach, we train a BiGAN with the proposed training approach to detect taxpayers, who are manipulating their tax returns.  For each taxpayer, we derive six correlation parameters and three ratio parameters from tax returns submitted by him/her. We train a BiGAN with the proposed training approach on this nine-dimensional derived ground-truth data set.  Next, we generate the latent representation of this data set using the  $encoder$ (encode this data set using the $encoder$) and regenerate this data set using the $generator$ (decode back using the $generator$) by giving this latent representation as the input. For each taxpayer, compute the cosine similarity between his/her ground-truth data and regenerated data. Taxpayers with lower cosine similarity measures are  potential return manipulators. We applied our method to analyze the iron and steel taxpayer’s data set provided by the Commercial Taxes Department, Government of Telangana, India. 
\end{abstract}


\section{Introduction}

\subsection{Outlier Detection}
Outliers are data points that are far from other data points. In other words, they are unusual values in a data set. Detecting outliers is essentially identifying unexpected elements in data sets \cite{chandola}. Outlier identification has several real world applications such as fraud detection and intrusion detection.  Outlier detection is a challenging activity. Several machine learning techniques are proposed in the literature for outlier detection.

Several methods derived from neural networks have been applied to outlier detection. Generative Adversarial Networks have emerged as a leading technique.  Goodfellow et al.  proposed  a new approach for training generative models via an adversarial process \cite{godfellow}. This method simultaneously trains two neural network modules: a generator module that captures the probability distribution of the training data and a discriminator module that estimates the probability that a sample came from the training data set rather than generated by the generator. In this way, the two modules are competing against each other, they are adversarial in the game theory sense, and are playing a zero-sum game. In their existing form, GANs have no means of learning the inverse mapping (projecting training data back into the latent space). Jeff Donahue et al.  proposed {\it Bidirectional Generative Adversarial Networks} (BiGANs) as a method of learning this inverse mapping \cite{jeff}.   Kaplan et al. implemented anomaly detection using BiGAN, considering it as a one-class anomaly detection algorithm \cite{KAPLAN}. Since generator and discriminator are highly dependent on each other in the training phase, to minimize this dependency, they proposed two different training approaches for BiGAN by adding extra training steps to it. They also demonstrated that the proposed approaches increased the performance of BiGAN on the anomaly detection tasks.

\subsection{Indirect Tax}
Indirect tax is collected by an intermediary (such as a retailer and manufacturer) from the consumer of goods or services \cite{dani}. The intermediary submits the tax he/she collected to the government by filing tax return forms at regular time intervals. In reality, the intermediary acts as a conduit for the flow of tax from the consumer of the goods or services to the government.

\subsubsection{Goods and Services tax}
Goods and Services Tax (GST) is a destination-based, multi-stage, comprehensive taxation system. In GST, the tax is levied in an incremental manner at each stage of the supply chain based on the value added to goods/services at that stage. This tax is levied at each stage of the supply chain  in such a way that the tax paid on purchases ($\it{Input\,tax}$ or $\it{input\,tax\,credit}$) is  given as a set-off for the tax levied on sales ($\it{output\,tax}$ or {\it liability}).

\begin{figure}[ht]
  \includegraphics[width=\linewidth]{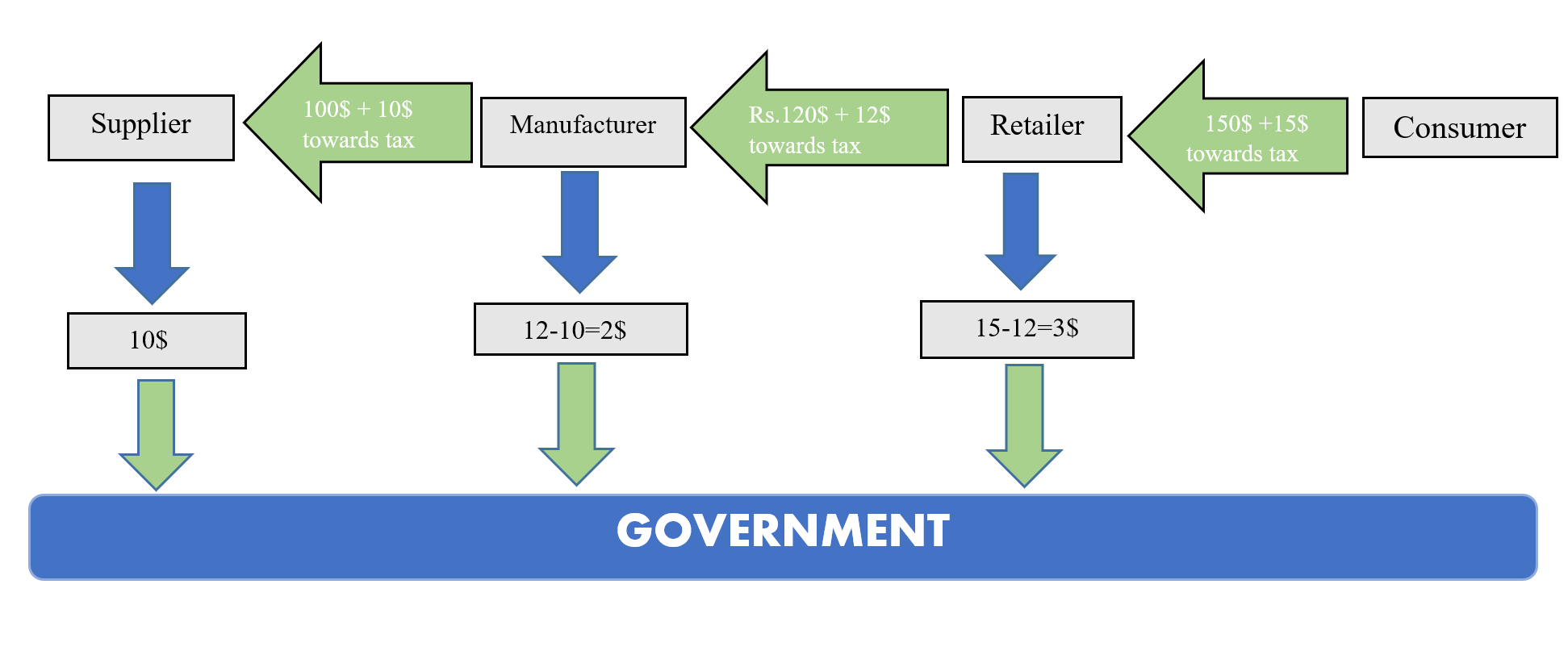}
  \caption{Tax Flow in GST}
  \label{fig:tax_flow}
\end{figure}

Figure \ref{fig:tax_flow} shows how  tax is collected incrementally at each stage of the supply chain. In this example, the manufacturer purchases goods from the supplier for a value of \$100 and pays \$10 as tax at a tax rate of 10\%. The supplier then pays the tax he/she collected to the government. In the next stage of the supply chain, the retailer purchases  finished goods from the manufacturer for a value of \$120 and pays \$12 as tax at a tax rate of 10\%. The  manufacturer pays $(\$12-\$10=\$2)$ to the government, which is the difference between the tax he paid to the supplier and  the tax he collected from the retailer. Finally, the consumer buys it from the retailer for a value of \$150 and pays \$15 as tax at a tax rate of 10\%. So the retailer will pay $(\$15-\$12=\$3)$ to the government. In essence, for every dealer in GST, {\it the tax payable is  the difference between output tax and input tax}.

\subsubsection{Tax evasion}
Taxation and tax evasion go hand in hand. It is a never-ending cat and mouse game. Business dealers manipulate their tax returns to avoid tax and maximize their profits. Tax enforcement officers formulate new rules and regulations to control tax evasion after studying the behavior of known tax evaders who exploit the loopholes in the existing taxation laws. In this game tax evaders always try their best  to stay a few steps ahead of the enforcement officers. Hence, it is necessary for the officials to identify the evasion as early as possible and close the loopholes before the techniques become a widespread practice. In this manner, the taxation officers will be able to limit the loss of government revenue due to tax evasion. The following are prominent ways of tax evasion.

\begin{enumerate}
\item The dealer will collect tax at a higher rate from the customer and pays it to the government at a lower rate.
\item The dealer does not report all the sales transactions made by her/him (sales suppression).
\item The dealer will show a lower taxable turnover by wrongly applying the prescribed calculations.
\item The dealer creates fictitious sales transactions where there is no movement of goods but only the invoices are circulated in order to claim an Input Tax Credit (ITC) and evade tax payment.  This method is called bill trading.

\end{enumerate}

\subsection{Our Contribution}

In this article, we propose a new training approach for bidirectional GAN (BiGAN) to detect outliers. This is an enhancement to the BiGAN training approach given in \cite{KAPLAN}. To validate the proposed approach, we train a BiGAN with the proposed training approach to detect taxpayers, who are manipulating their tax returns in the Goods and Services Taxation system, which came into operation in India in July $2017$.

The Goods and Services Taxation system unified the taxation laws in India. As per this system, dealers are supposed to file tax return statements every month by providing the complete details of sales and purchases that happened in the corresponding month. The objective of this work is to  identify  dealers who manipulate their tax return statements to minimize their tax liability. We train {\it Bidirectional GAN} (BiGAN) using the proposed approach on nine-dimensional ground-truth data derived from tax returns submitted by taxpayers (six correlation parameters and three ratio parameters). Next, we encode the ground-truth data set using the  $encoder$ (generate latent representation) and decode it back using the $generator$ (regenerate the ground-truth data) by giving the latent representation as input. For each taxpayer, compute the cosine similarity between his/her ground-truth data and regenerated data. Taxpayers with lower cosine similarity measures are  potential return manipulators. This idea can be applied in other nations where multi-stage indirect taxation is followed.

The rest of the paper is organized as follows. In Section \ref{pre}, we discuss the previous relevant works. In Section \ref{daused}, we will explain the data set used. In Section \ref{ebgan}, we explain our proposed enhancement for BiGAN.  In Section \ref{metho}, we give a detailed description of the methodology used in this paper. The results obtained are discussed in Section \ref{exper}. 

\section{Related Work}
\label{pre}

Chandola et al. presented several data mining techniques for anomaly detection \cite{chandola}.    Daniel de Roux et al. presented a very interesting approach for the detection of fraudulent taxpayers using only unsupervised learning methods \cite{daniel}.   Yusuf Sahin et al.  worked on credit card fraud detection \cite{yusuf}. They developed some classification models based on Artificial Neural Networks (ANN) and Logistic Regression (LR). This study is one of the first in credit card fraud detection with a real data set to compare  the performance of ANN and LR. Zhenisbek Assylbekov et al.  presented statistical techniques for detecting VAT evasion by Kazakhstani business firms \cite{zhen}. Starting from features selection they performed an initial exploratory data analysis using Kohonen self-organizing maps.  Hussein et al.  described classification-based and clustering-based anomaly detection techniques \cite{hussein}. They applied {\it K-Means} to a refund transaction data set from a telecommunication company, with the intent of identifying fraudulent refunds.  Gonzlez et al.  described methods to detect potential false invoice issuers/users based on the information in their tax return statements using different types of data mining techniques \cite{Gonzlez}. First, clustering algorithms like SOM and neural gas are used to cluster similar taxpayers. Then decision trees, neural networks, and Bayesian networks are used to identify those features that are related to the conducting of fraud and/or no fraud. Song Wang et al. introduced the challenges of anomaly detection in the traditional network, as well as in the next generation network, and reviewed the implementation of machine learning in the anomaly detection under different network contexts \cite{wang}. Shuhan Yuan et al. used a deep learning approach for fraud detection \cite{yuan}. This method will work only for the labeled data set. Jian Chen et al. applied deep learning techniques to credit card fraud detection. The first used sparse autoencoder to obtain representations of normal transactions and then trained a generative adversarial network (GAN) with these representations. Finally, they combined the SAE and the discriminator of GAN and apply them to detect whether a transaction is genuine or fraud \cite{Chen}. Shenggang Zhang  et al. proposed an anomaly detection model based on BiGAN for software defect prediction. The model proposed by them not only does not need to consider the class imbalance problem but also uses a semi-supervised method to train the model \cite{zhang}. Raghavendra Chalapathy et al. presented a structured and comprehensive overview of research methods in deep learning-based anomaly detection. Furthermore, they reviewed the adoption of these methods for anomaly across various application domains and assess their effectiveness \cite{raghav}.

\section{Description of the Data Set}
\label{daused}
\begin{figure}[!tbp]
  \centering
  \begin{minipage}[b]{0.45\textwidth}
    \includegraphics[width=\textwidth]{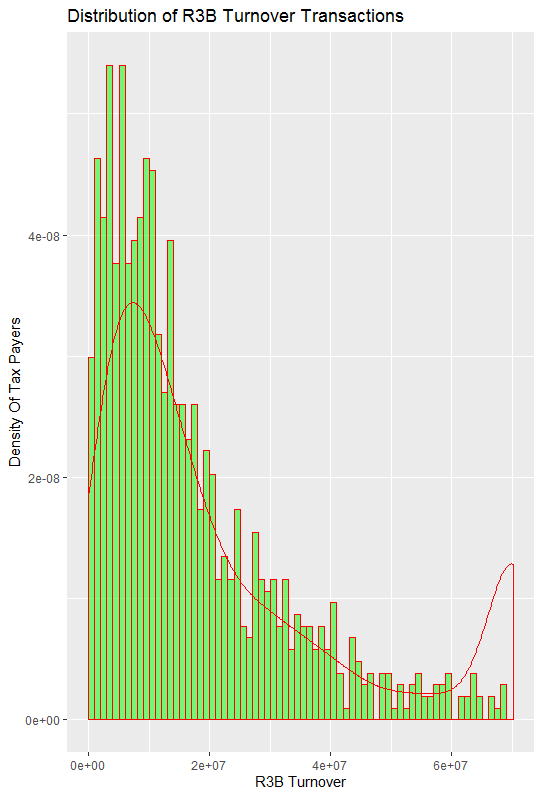}
    \caption{Distribution of Turnover}
    \label{R3B_TO}
  \end{minipage}
  \hfill
  \begin{minipage}[b]{0.45\textwidth}
    \includegraphics[width=\textwidth]{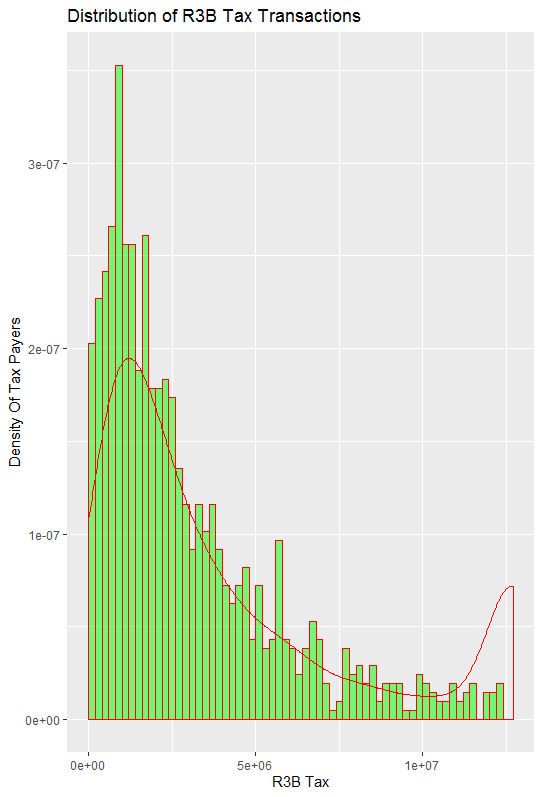}
    \caption{Distribution of Liability}
    \label{R3B_Tax}
  \end{minipage}
\end{figure}

\begin{figure}[!tbp]
  \centering
  \begin{minipage}[b]{0.45\textwidth}
    \includegraphics[width=\textwidth]{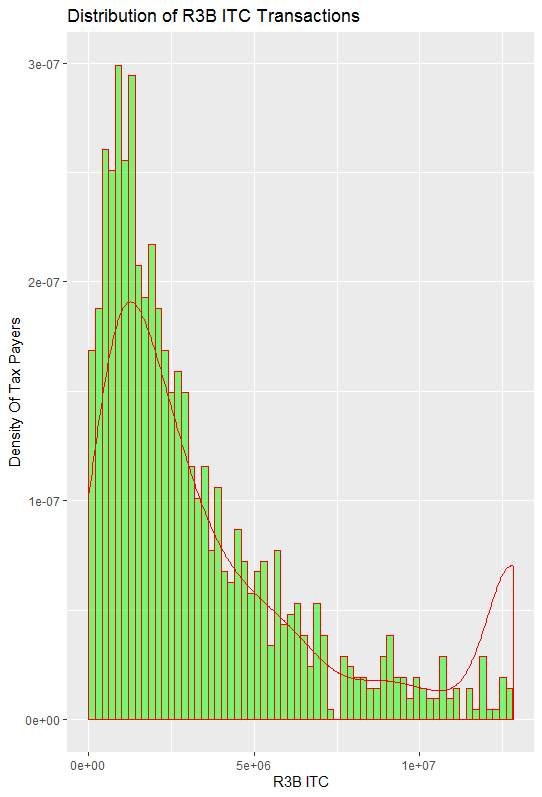}
    \caption{Distribution of Input Tax Credit}
    \label{R3B_ITC}
  \end{minipage}
  \hfill
  \begin{minipage}[b]{0.45\textwidth}
    \includegraphics[width=\textwidth]{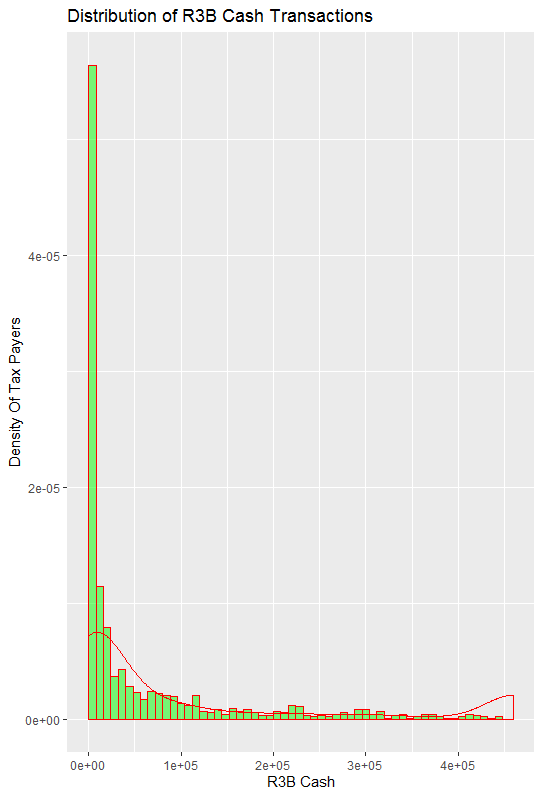}
    \caption{Distribution of Cash Payments}
    \label{R3b_cash}
  \end{minipage}
\end{figure}

GSTR-3B is a monthly return that has to be filed by the dealer. It is a simple return in which a summary of Input Tax Credit along with outward supplies are declared, and payment of tax is affected by the taxpayer. Table \ref{taxdetials} is a sample of  GST  returns data. Each row in this table corresponds to a monthly return by a dealer. {\it ITC (Input tax credit)} is the amount of tax paid during purchases of services/goods by the dealer. The {\it output tax} is the amount of tax  collected by the dealer during the sales of services/goods. The dealer has to pay the Government the gap between the  {\it output tax} and {\it ITC}, i.e., output tax - ITC.  The actual database consists of much more information, like,  return filing data, tax payment method, exempted sales, international exports, and sales on RCM (reverse charge mechanism). Figures \ref{R3B_TO}, \ref{R3B_Tax},  \ref{R3B_ITC}, and  \ref{R3b_cash}  show the distribution of turnover, liability, input tax credit, and cash payments.


\begin{table}[ht]
\begin{center}
\begin{tabular}{|c|c|c|c|c|c|c|}
\hline
S.No.&Firm &Month & Purchases & Sales & ITC & Output Tax\\
\hline
1 &BC& Jan 2019 & 190000  & 210000 & 20200 & 24000\\
\hline
2 &BE& Sep 2021 & 202000  & 270000  & 5200 & 9200\\
\hline
3 &BD& Oct 2021&  400200  & 420000  & 41000 & 43000\\
\hline
\end{tabular}
\end{center}
\caption{GST Returns Data (GSTR-3B)}
\label{taxdetials}
\end{table}


\section{Enhanced BiGAN}
\label{ebgan}
\subsection{Generative Adversarial Network}
A generative adversarial network (GAN) is a recent invention in  deep learning designed by Ian Goodfellow and his colleagues \cite{godfellow}. Two neural network modules (generator, discriminator) compete with each other in a simultaneous game. Given a training data set (ground-truth data set), the generator learns to generate new data with the same statistics as the training data set and the discriminator learns to estimate the probability that the given input came from the training data set rather than generated by the generator.

The discriminator's parameters are optimized to minimize the loss function in Equation \ref{do}. Here $DI$ is the estimated probability by the discriminator that input came from the ground-truth data set rather than generated by the generator given the  ground-truth data set as input, and $DG$ is the estimated probability  by the discriminator that input came from the ground-truth data set rather than generated by the generator given  the  generated  data set as input. The generator's parameters are optimized  to minimize the loss function in  Equation \ref{dg}. 

\begin{equation}
\label{do}
    -1*(log(DI)+log(1-DG))
\end{equation}
\begin{equation}
\label{dg}
    -1*(log(DG))
\end{equation}

\subsection{Bidirectional GAN}
Bidirectional GAN (BiGAN) is a representation learning method. BiGAN adds an encoder to the standard  GAN architecture.
The encoder takes ground-truth data set $X$ and outputs a latent representation $E(X)$ of this data set. The generator takes a random sample $z$ and generates a data set $G(z)$.  The BiGAN discriminator discriminates not only $X$ versus $G(z)$, but jointly in data and latent space ( discriminates  tuple (X,E(X)) versus tuple (G(z),z)).

The discriminator is trained to minimize Equation \ref{bd}, where $DE$ is the estimated probability by the discriminator that input came from $(X,E(X))$ rather than $(G(z),z)$ given $(X,E(X))$ as the input, and $DG$ is the estimated probability by the discriminator that input came from $(X,E(X))$ rather than $(G(z),z)$ given $(G(z),z)$ as the input. The generator is trained to minimize Equation \ref{bg}. The encoder is trained to minimize  Equation \ref{be}

\begin{equation}
\label{bd}
    -1*(log(DE)+log(1-DG))
\end{equation}

\begin{equation}
\label{bg}
    -1*(log(DG))
\end{equation}

\begin{equation}
\label{be}
    -1*log(1-DE)
\end{equation}

\subsection{Enhanced Bidirectional GAN}

Kaplan et al. suggested one more cost function to optimize generator and encoder parameters of BiGAN \cite{KAPLAN}. Let $X$ be the current batch of input data set and $E(X)$ be the output of $encoder$ with $X$ as input. Let $G(E(X))$ be the output of $generator$ with $E(X)$ as input. They suggested to update $encoder's$ and $generator's$ parameters together to  minimize the mean of euclidean distance between $X$ and $G(E(X))$ as in Equation \ref{kapeq}.

\begin{equation}
\label{kapeq}
    mean~of~euclidean~distance(X,G(E(X)))
\end{equation}

\begin{figure}[!tbp]
  \centering
    \includegraphics[width=0.4\textwidth]{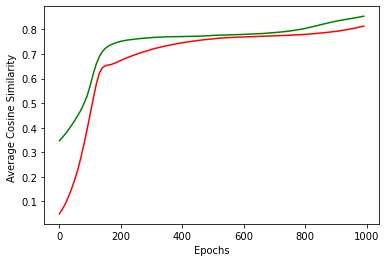}
    \caption{Epochs Vs Cosine Similarity Measure}
    \label{simcomp}
\end{figure}

Cosine similarity is a measure used to find similarity between two vectors. Mathematically, it measures the cosine of the angle between two vectors projected in a multi-dimensional space. We propose that increasing the cosine similarity between $X$ and $G(E(X)$ is a better approach than minimizing the euclidean distance between them. We update $encoder's$ and $generator's$ parameters together to  increase the value of  Equation \ref{kapneq}. Figure \ref{simcomp} shows cosine similarity between ground-truth data and regenerated data for a different number of epochs. The red coloured curve shows the result of the algorithm in \cite{KAPLAN} and the green coloured curve shows the result of the proposed method.   Algorithm \ref{alg:one} gives a detailed description of the enhanced BiGAN training procedure. 

\begin{equation}
\label{kapneq}
    mean~of~cosine~similarity(X,G(E(X)))
\end{equation}

\SetKwComment{Comment}{/* }{ */}
\RestyleAlgo{ruled}
\LinesNumbered
\begin{algorithm}
\DontPrintSemicolon
\SetAlgoLined
\SetNoFillComment
\caption{Proposed BiGAN Training}\label{alg:one}
\KwData{Nine-Dimensional Ground-Truth Data}
\KwResult{Trained BiGAN}

\For{  number of training epochs}{
    \For{number of batches}{
    \tcc{Training Discriminator}
    Let $z$ be random sample from a standard normal distribution \;
    $G(z) \gets$ output of $generator$ with $z$ as input\;
    Let $X$ be the current batch of input data set\;
    $E(X) \gets$ output of $encoder$ with $X$ as input\;
    $DE \gets$ output of $discriminator$ with $(X,E(X))$ as input\;
    $DG \gets$ output of $discriminator$ with $(G(z),z)$ as input\;
    Update $discriminator's$ parameters to maximize $log(DE)+log(1-DG)$\;
    \;
    \tcc{Training Generator}
    Let $z$ be random sample from a standard normal distribution \;
    $G(z) \gets$ output of $generator$ with $z$ as input\;
    $DG \gets$ output of $discriminator$ with $(G(z),z)$ as input\;
    Update $generator's$ parameters to maximize $log(DG)$\;
    \;
    \tcc{Training Encoder}
    Let $X$ be the current batch of input data set\;
    $E(X) \gets$ output of $encoder$ with $X$ as input\;
    $DE \gets$ output of $discriminator$ with $(X,E(X))$ as input\;
    Update $encoder's$ parameters to maximize $log(1-DE)$\;
    \;
    \tcc{Reducing Mismatch Between Encoder and Generator}
    Let $X$ be the current batch of input data set\;
    $E(X) \gets$ output of $encoder$ with $X$ as input\;
    $G(E(X)) \gets$ output of $generator$ with $E(X)$ as input\;
    Update $encoder's$ and $generator's$ parameters together to  increase the cosine similarity  between $X$ and $G(E(X))$\;
    }
}
\end{algorithm}

\section{Methodology}
\label{metho}
The objective of this work is to train Bidirectional GAN (BiGAN) using the proposed training approach on nine-dimensional ground-truth data derived from tax returns submitted by taxpayers (six correlation parameters and three ratio parameters) to  identify malicious dealers who manipulate their tax return statements. We had taken the data sets explained in Section \ref{daused} from July 2017 to March 2022 and derived nine features (parameters) from these data sets.
\begin{itemize}
\item Six are sensitive correlation parameters.
\item Three are ratio parameters.
\end{itemize}

In Subsection \ref{cor}, we explain the six correlation parameters that are used. In Subsection \ref{rat}, we describe the three ratio parameters that are used.  In Subsection \ref{bgan}, we give a detailed algorithm. 

\subsection{Correlation  parameters}
\label{cor}
In the Indian GST system,   three types of  taxes are collected, $viz.,$ CGST, SGST, and IGST.
\begin{itemize}
\item{\it CGST:} Central Goods and Services Tax is levied on intrastate transactions and collected by the Central Government of India.
\item{\it SGST:} State/Union Territory Goods and Services Tax, which is also levied on intrastate transactions and collected by the state or union territory Government.
\item{\it IGST:} Integrated Goods and Services Tax is levied on interstate sales.  Central Government takes half of this amount and passes the rest of the amount to the state, where corresponding goods or services are consumed.
\end{itemize}

 The six  correlation parameters are mentioned in Table \ref{cp}.  Total GST liability is the sum of CGST liability, SGST liability, and IGST liability. Total ITC is equal to the sum of SGST ITC, CGST ITC, and IGST ITC.


\begin{table}
\begin{center}
\begin{tabular}{|c|l|}

\hline
S. No. & \hspace{0.25in} The Six Correlation Parameters \\
\hline
1  &  Total GST Liability {\t VS} Total Sales Amount\\
\hline
2 &  SGST Liability {\it VS}  Total GST Liability\\
\hline
3  &  SGST paid in cash {\it VS} SGST Liability \\
\hline
4 &  SGST paid in cash {\it VS} Total Sales Amount\\
\hline
5 &   Total ITC {\it VS} Total Tax Liability\\
\hline
6 &  IGST ITC {\it VS} Total ITC\\
\hline
\end{tabular}
\end{center}
\caption{correlation parameters }
\label{cp}
\end{table}

\subsection{Ratio parameters}
\label{rat}
\begin{enumerate}
\item{The ratio of $Total\,Sales$ VS. $Total\,Purchases$}: This ratio captures the value addition.\\
\item{The ratio of $IGST\,ITC$ VS. $Total\,ITC$}: This ratio captures how much purchase is shown as interstate or imports compared to total purchases.\\
\item{The ratio of $Total\,Tax\,Liability$ VS. $IGST\,ITC$}.

\end{enumerate}

\subsection{Identifying Fraudulent Taxpayers  using BiGAN}
\label{bgan}

\begin{algorithm}
\DontPrintSemicolon
\SetAlgoLined
\SetNoFillComment
\caption{Identifying Fraudulent Taxpayers}\label{alg:two}
\KwData{Nine Dimensional Ground-Truth Data (GT)}
\KwResult{Fraudulent Taxpayers}
    Train a BiGAN with the ground-truth data set $GT$ using  Algorithm \ref{alg:one}\;
    $LR \gets$  latent representation of the ground-truth data set $GT$ computed using the $encoder$\;
    $RG \gets $ output of the $generator$ with $LR$ as the input \;
    $CS \gets $ Cosine similarity between corresponding rows of $GT$ and $RG$\;
    $Fraud set \gets $ taxpayers whose similarity score is less than $first~quantile - 1.5*IQR$\; 
\end{algorithm}

\section{Experimentation and Results obtained}
\label{exper}

\begin{figure}[!tbp]
  \centering
  \begin{minipage}[b]{0.3\textwidth}
    \fbox{\includegraphics[width=\textwidth]{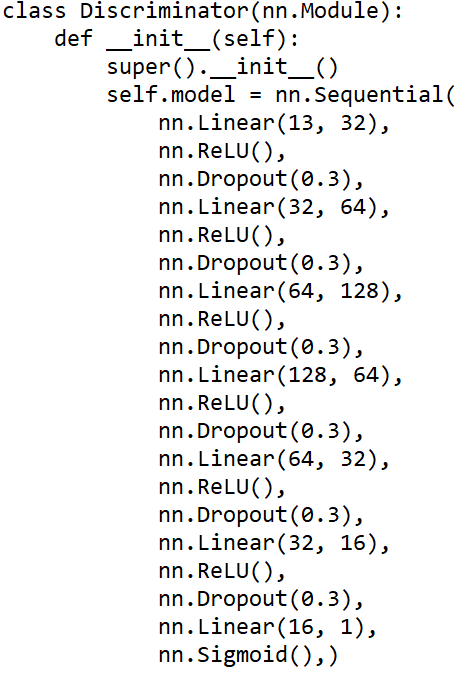}}
    \caption{Discriminator}
    \label{discriminator}
  \end{minipage}
  \hfill
  \begin{minipage}[b]{0.3\textwidth}
    \fbox{\includegraphics[width=\textwidth]{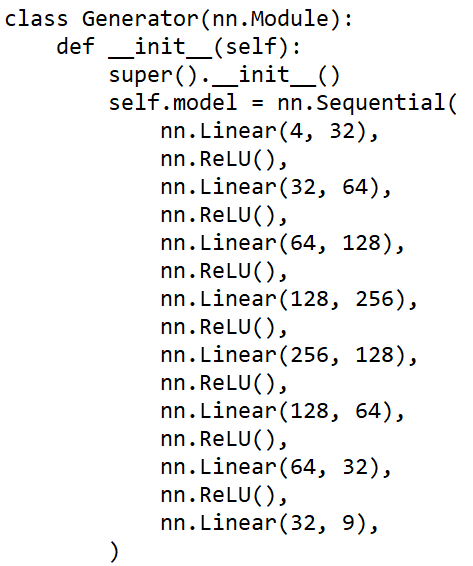}}
    \caption{Generator}
    \label{generator}
  \end{minipage}
  \hfill
  \begin{minipage}[b]{0.3\textwidth}
    \fbox{\includegraphics[width=\textwidth]{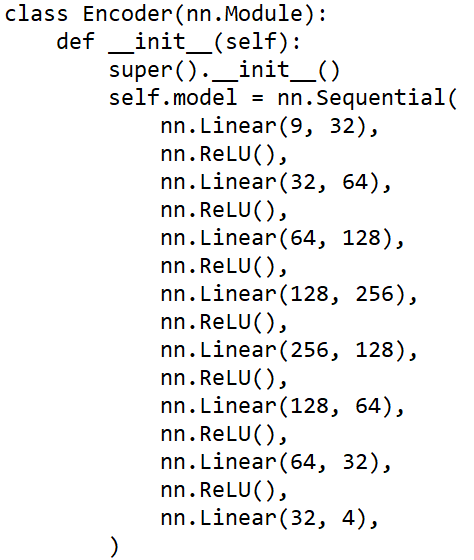}}
    \caption{Encoder}
    \label{encoder}
  \end{minipage}
\end{figure}

\begin{figure}[!tbp]
  \centering
  \begin{minipage}[b]{0.4\textwidth}
    \includegraphics[width=\textwidth]{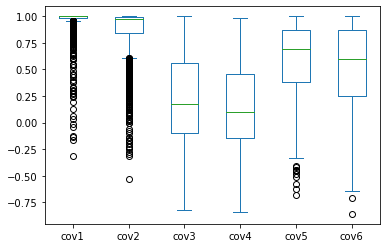}
    \caption{Correlation Parameters}
    \label{corpar}
  \end{minipage}
  \begin{minipage}[b]{0.4\textwidth}
    \includegraphics[width=\textwidth]{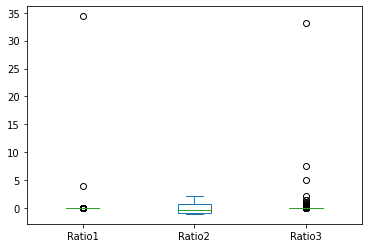}
    \caption{Ratio Parameters}
    \label{ratiopar}
  \end{minipage}
\end{figure}

We had taken returns data of 1184 iron and steel dealers. We computed correlation and ratio parameters defined in subsections \ref{cor} and \ref{rat} for each dealer.  Table \ref{table1} gives a snapshot of the parameters created for each dealer. Figure \ref{corpar} and Figure \ref{ratiopar} show the distribution of correlation and ratio parameters respectively.  Figures \ref{discriminator}, \ref{generator}, and \ref{encoder} give the PyTorch code of discriminator, generator, and encoder respectively. After experimentation, we opted for four-dimensional latent space. Note that input to the discriminator is thirteen dimensions (four-dimensional latent space data and  nine-dimensional training/ground-truth data set). 

\begin{figure}[!tbp]
    \includegraphics[width=0.4\textwidth]{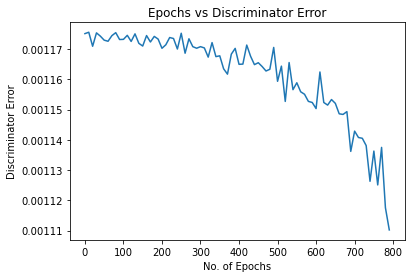}
    \caption{Discriminator Error}
    \label{discrimierror}
\end{figure}

\begin{figure}[!tbp]
  \centering
  \begin{minipage}[b]{0.4\textwidth}
    \includegraphics[width=\textwidth]{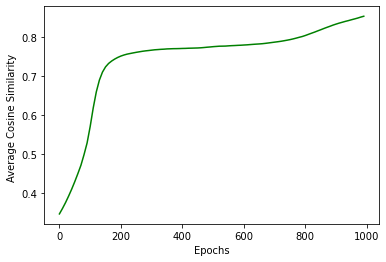}
    \caption{Avg Cosine Measure}
    \label{cosineavg}
  \end{minipage}
  \begin{minipage}[b]{0.4\textwidth}
    \includegraphics[width=\textwidth]{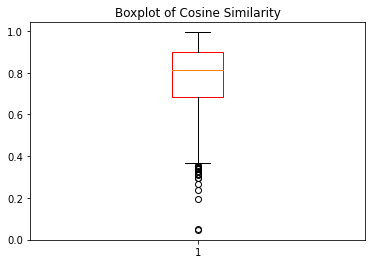}
    \caption{Final Cosine  Measures}
    \label{cosinebox}
  \end{minipage}
\end{figure}

Figure \ref{discrimierror} shows the discriminator's error at different epochs. Figure \ref{cosineavg} shows the average cosine similarity measure between corresponding rows in the training data set and regenerated data set at different epochs. The boxplot in  Figure \ref{cosinebox} shows the distribution of cosine similarities between corresponding rows in the training data set and regenerated set at the final epoch. The third quantile value is  0.8992, first quantile value is 0.6827. There are nineteen taxpayers whose cosine similarity values are less than the $first~quantile - 1.5*IQR$. Tax returns of these taxpayers need further investigation by tax officers. Expected evasion by these taxpayers is more than a few hundred million  Indian rupees.    

\begin{table}[ht]
\begin{adjustbox}{width=\columnwidth,center}
\centering
\begin{tabular}{|c|c|c|c|c|c|c|c|c|c|}
\hline
\begin{tabular}[c]{@{}c@{}}S.No\end{tabular} & Corr 1   & Corr 2   & Corr 3    & Corr 4    & Corr 5    & Corr 6   & \begin{tabular}[c]{@{}c@{}}Total Sales\\ /Total Purchases\end{tabular} & \begin{tabular}[c]{@{}c@{}}IGST ITC\\ /Total ITC\end{tabular} & \begin{tabular}[c]{@{}c@{}}Total tax liability\\ /IGST ITC\end{tabular}  \\ \hline
1                                                    & 0.9977 & 0.9998 & 0.2159  & 0.1967  & 0.9556  & 0.9988 & 1.0465                                                                 & 0.8717                                                        & 1.3272                                                                             \\ \hline
2                                                    & 0.9940 & 0.9799 & -0.3371 & -0.2486 & 0.6408  & 0.5539 & 1.1992                                                                 & 0.1347                                                        & 7.4129                                                                             \\ \hline
3                                                    & 0.9476 & 0.4556 & 0.0017  & 0.1286  & -0.1620 & 0.9606 & 1.6991                                                                 & 0.8020                                                        & 2.1824                                                                             \\ \hline
\end{tabular}
\end{adjustbox}
\vspace{0.05in}
\caption{Snapshot of parameters}
\label{table1}
\end{table}

\begin{table}[ht]
\begin{adjustbox}{width=\columnwidth,center}
\centering
\begin{tabular}{|c|c|c|c|c|c|c|c|c|c|}
\hline
\begin{tabular}[c]{@{}c@{}}S.No\end{tabular} & Corr 1   & Corr 2   & Corr 3    & Corr 4    & Corr 5    & Corr 6   & \begin{tabular}[c]{@{}c@{}}Total Sales\\ /Total Purchases\end{tabular} & \begin{tabular}[c]{@{}c@{}}IGST ITC\\ /Total ITC\end{tabular} & \begin{tabular}[c]{@{}c@{}}Total tax liability\\ /IGST ITC\end{tabular}  \\ \hline
1  & 0.39657727 &  0.91993311 &  0.49645071 &  0.57682353 &  0.80236321 & 0.67371039 & -0.03205805 &  0.44915969 & -0.05412638 \\ \hline
2 & 0.99616234 &   0.89398749 &  0.7674135  &  0.60683971 &  0.97080785 & 0.88260176 & -0.03226673 &  0.36960728 & -0.05415696 \\ \hline
3 & 0.99947796 &   0.98932055 &  0.65413855 &  0.70517677 &  0.35052473 & 0.87797351 & -0.03098378 &  0.60539265 & -0.05411582 \\\hline
\end{tabular}
\end{adjustbox}
\vspace{0.05in}
\caption{Snapshot of normalized parameters of few genuine dealers}
\label{caseg}
\end{table}

\begin{table}[ht]
\begin{adjustbox}{width=\columnwidth,center}
\centering
\begin{tabular}{|c|c|c|c|c|c|c|c|c|c|}
\hline
\begin{tabular}[c]{@{}c@{}}S.No\end{tabular} & Corr 1   & Corr 2   & Corr 3    & Corr 4    & Corr 5    & Corr 6   & \begin{tabular}[c]{@{}c@{}}Total Sales\\ /Total Purchases\end{tabular} & \begin{tabular}[c]{@{}c@{}}IGST ITC\\ /Total ITC\end{tabular} & \begin{tabular}[c]{@{}c@{}}Total tax liability\\ /IGST ITC\end{tabular}  \\ \hline
1 &  0.31020196 &  0.71482671 &  0.9993972 &  0.45058559 &  0.95149288 & 0.99453044 &  3.95717384 & -0.56773328 & -0.04284052 \\ \hline
2 &  0.99998812 &  0.99800189 & -0.26364705 & -0.27946686 &  0.49037541 & 0.35609805 &  -0.03317666 & -1.06640099 &  7.48629588 \\ \hline
3 & 0.0335644 &  -0.1469533  & -0.14223751 & -0.21525574 &  0.65303945 & 0.54761113 & -0.02922851 & -0.84931729 & -0.05217054 \\ \hline
\end{tabular}
\end{adjustbox}
\vspace{0.05in}
\caption{Snapshot of normalized parameters of few fraudulent dealers}
\label{casef}
\end{table}

\subsection{Case Study}
\begin{figure}[!tbp]
\includegraphics[width=0.5\textwidth]{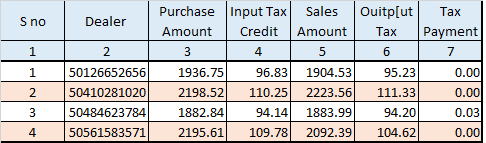}
\caption{Business Details}
\label{bdd}
 
\end{figure}

 Table \ref{caseg} gives normalized features of three genuine dealers. Table \ref{casef} gives normalized features of three fraudulent dealers. We observed that the feature {\it IGST ITC /total ITC} is positive for most of the genuine dealers and negative for most of the fraudulent dealers. This means fraudulent dealers are showing most of their purchases as intra-state purchases.  The fourth correlation parameter (total sales vs SGST paid in cash) is low for fraudulent dealers and high for genuine dealers. This means fraudulent dealers are not paying any cash and using the ITC to set off the liability. Figure \ref{bdd} shows the business details of a few fraudulent taxpayers.  Amounts are in lakhs of Indian Rupees. We can observe that they are not paying cash even though they are doing huge business.

\section{Conclusion}
\label{con}
In this paper, we enhanced the BiGAN training approach given in \cite{KAPLAN}.  This training approach significantly improved the performance/stability of BiGAN.

We analyzed the tax returns data set of a set of business dealers in the state of Telangana, India, to identify dealers who perform extensive tax evasion.  We had taken data of 1184 iron and steel dealers and derived ground-truth data set based on their monthly returns. The ground-truth data contains nine parameters (six are correlation parameters and three are ratio parameters). We used the proposed method of BiGAN  training to identify fraudulent dealers. First, we trained a BiGAN using the ground-truth data set. Next, we encoded  the ground-truth data set using the  $encoder$  and decoded it back using the $generator$ by giving the latent representation as input. For each taxpayer, we computed the cosine similarity between his/her ground-truth data and regenerated data. Taxpayers with lower cosine similarity measures are  potential return manipulators. We identified nineteen dealers whose cosine similarity score is  less than the $first~quantile - 1.5*IQR$. Expected evasion by these taxpayers is more than a few hundred million Indian rupees.

\section*{Acknowledgment}

We express our sincere gratitude to the Telangana state Government, India, for sharing the commercial tax data set, which is used in this work.

\bibliographystyle{IEEEtran}
\bibliography{main.bib}

\end{document}